\documentclass{article}
\usepackage{ijcai21}

\usepackage{times}

\usepackage{soul}
\usepackage{url}
\usepackage[hidelinks]{hyperref}
\usepackage[utf8]{inputenc}
\usepackage[small]{caption}
\usepackage{graphicx}
\usepackage{amsmath}
\usepackage{booktabs}

\usepackage{tikz}
\usepackage{tikz-qtree}
\usetikzlibrary{bayesnet}
\usetikzlibrary{trees,shapes,arrows}

\usepackage{bm}
\usepackage[lined,commentsnumbered]{algorithm2e}
\usepackage{wrapfig}
\usepackage{url}
\usepackage{amsthm}
\usepackage{chngcntr}

\usepackage{tcolorbox}
\usepackage{IEEEtrantools}
\usepackage{mathtools}  %
\usepackage{amssymb,amsmath}%

\usepackage[colorinlistoftodos]{todonotes}
\renewcommand{\vec}{\boldsymbol}
\newcommand{\pprob}{\operatorname{p}\probarg}
\DeclarePairedDelimiterX{\probarg}[1]{(}{)}{%
  \ifnum\currentgrouptype=16 \else\begingroup\fi
  \activatebar#1
  \ifnum\currentgrouptype=16 \else\endgroup\fi
}

\DeclarePairedDelimiterX{\probargq}[1]{(}{)}{%
  \ifnum\currentgrouptype=16 \else\begingroup\fi
  \activatebar#1
  \ifnum\currentgrouptype=16 \else\endgroup\fi
}
\newcommand{\innermid}{\nonscript\;\delimsize\vert\nonscript\;}
\newcommand{\activatebar}{%
  \begingroup\lccode`\~=`\|
  \lowercase{\endgroup\let~}\innermid
  \mathcode`|=\string"8000
}
\newcommand{\matr}[1]{\mathbf{#1}}
\RequirePackage{xcolor}

\newcommand{\exptt}[2]{\mathbb{E}_{#1}\left[#2\right]}
\newcommand{\eqdef}{\vcentcolon=}
\newcommand{\kl}[2]{\mathbb{KL}\left[{#1}\parallel#2\right]}

\newcommand{\softmax}[1]{\text{softmax}\left(#1\right)}

\newcommand{\aside}[1]{}

\title{Topic Modelling Meets Deep Neural Networks: A Survey}

\author{
He Zhao\and
Dinh Phung\and
Viet Huynh\and
Yuan Jin\and
Lan Du\and
Wray Buntine
\\
\affiliations
Department of Data Science and Artificial Intelligence, Monash University, Australia
\emails
\{ethan.zhao, dinh.phung, viet.huynh, yuan.jin, lan.du, wray.buntine\}@monash.edu
}

\begin{document}

\maketitle

\begin{abstract}
\aside{Topic modelling has been successfully used as a powerful technique for text analysis for almost twenty years. When topic modelling meets deep neural networks in the deep learning era, there emerges a new and increasingly popular research area, \textit{neural topic models}, with a hundred models developed and a wide range of applications in neural language understanding such as language models, text generation and summarisation.
It urgently calls to summarise the research developments and discuss open problems and future directions.
In this paper, we provide a focused yet comprehensive overview of neural topic models for interested researchers in the AI community, so as to facilitate them to navigate and innovate in this fast-growing research area. To the best of our knowledge, ours is the first review focusing on this specific topic.}
Topic modelling has been a successful technique for text analysis for almost twenty years. When topic modelling met deep neural networks, there emerged a new and increasingly popular research area, \textit{neural topic models}, with over a hundred models developed and a wide range of applications in neural language understanding such as text generation, summarisation and language models.
There is a need to summarise research developments and discuss open problems and future directions.
In this paper, we provide a focused yet comprehensive overview of neural topic models for interested researchers in the AI community, so as to facilitate them to navigate and innovate in this fast-growing research area. To the best of our knowledge, ours is the first review focusing on this specific topic.
\end{abstract}

\section{Introduction}

A powerful technique for text analysis, topic modelling has enjoyed success in various applications in machine learning, natural language processing (NLP), and data mining for almost two decades. 
A topic model is applied to a collection of documents and aims to discover a set of latent topics, each of which describes an interpretable semantic concept.
Bayesian probabilistic topic models (BPTMs) have been the most popular and successful series of models, with latent Dirichlet allocation (LDA) the best known representative. A BPTM usually specifies a probabilistic generative model that generates the data of a document with a structure of latent variables sampled from pre-specified distributions connected by Bayes' theorem. Topics are captured by these latent variables.
Like other Bayesian models, the learning of a BPTM is done by a (Bayesian) inference process (e.g.~variational inference (VI) and Monte Carlo Markov Chain sampling).

Despite their success, conventional BPTMs started to show signs of fatigue in the era of big data and deep learning: 
\textbf{1)} Given a specific BPTM, its inference process usually needs to be customised accordingly and the inference complexity may grow significantly as the model complexity grows. Unfortunately, it is also hard to automate the design of the inference processes.
\textbf{2)} The inference processes for conventional BPTMs can be hard to scale efficiently on large text collections or to leverage parallel computing facilities like GPUs.
\textbf{3)} It is usually inconvenient to integrate BPTMs with other deep neural networks (DNNs) for joint training.

With the recent developments in DNNs and deep generative models, there has been an emerging research direction which aims to leverage DNNs to boost performance, efficiency, and usability of topic modelling, named \textit{neural topic models} (NTMs).
With appealing flexibility and scalability, NTMs have gained a huge research following, with more than a hundred models and variants developed to date. Moreover, NTMs have been used in important NLP tasks including text generation, document summarisation, and translation,  areas to which conventional topic models are harder to apply.
Therefore, it is important to properly summarise research developments, categorise existing approaches, identify remaining issues, and discuss open problems and future directions.
To the best of our knowledge, a comprehensive review specifically focusing on NTMs has not been published. In this paper, we would like to fill this gap by providing an overview for interested researchers who want to develop new NTMs and/or to apply NTMs in their domains. The notable contributions of our paper can be summarised as follows:
\textbf{1)} We propose a taxonomy of NTMs where we categorise existing models based on their backbone framework.
\textbf{2)} We first provide an informative discussion and overview of the background and evaluation methods for NTMs and conduct a focused yet comprehensive review, offering detailed comparisons of the variants in different categories of NTMs with applications.
\textbf{3)} We identify the limitations of existing methods and analyse the possible future research directions for NTMs.

The rest of this paper is organised as follows. Section~\ref{sec-background} introduces the background, definitions, and evaluations. Section~\ref{sec-vae} and~\ref{sec-other-framework} review NTMs with various backbone frameworks. Section~\ref{sec-application} discusses the applications. The current challenges and future directions are discussed in Section~\ref{sec-discuss}. 
Note that given the two page limit of references, we may only keep the most relevant papers to NTMs.

\section{Background, Definition, and Evaluation}
\label{sec-background}
\subsection{Background and Definition}
The most important idea of a topic model is the modelling of the three key entities: \textit{document}, \textit{word}, and \textit{topic}.

\paragraph{Notations of Data} A topic model works on a corpus (i.e., a collection of documents), where a document, by its nature, can be represented as a sequence of words, which can be denoted by a vector of natural numbers, $\vec{s} \in \mathbb{N}^L$, where $L$ is the length of the document and $s_j \in \{1,\cdots,V\}$ is the index in the vocabulary (with the size of $V$) of the token for the $j^{\text{th}}~(j \in \{1,\cdots,L\})$ word.
A more common representation in topic modelling is the bag-of-words model, which represents a document by a vector of word counts, $\vec{b} \in \mathbb{Z}_{\ge 0}^V$, where $b_v$ indicates the occurrences of the vocabulary token $v \in \{1,\cdots, V\}$ in the document. One can readily obtain $\vec{b}$ for a document from its word sequence vector $\vec{s}$.

\paragraph{Notations of Latent Variables} A central concept is a \textit{topic}, which is usually interpreted as a cluster of words, describing a specific semantic meaning. A topic is or can be normalised into a distribution over the tokens in the vocabulary, named \textit{word distribution}, $\vec{t} \in \Delta^V$, where $\Delta^V$ is a $V$ dimensional simplex and $t_v$ indicates the weight or relevance of token $v$ under this topic. Usually, a document's semantic content is assumed to be captured or generated by one or more topics shared across the corpus. Therefore, a document is commonly associated with a distribution (or a vector that can be normalised into a distribution) over $K$ ($K \ge 1$) topics, named \textit{topic distribution}, $\vec{z} \in \Delta^K$, where $z_k$ indicates the weight of the $k^{\text{th}}$ topic for this document. We further use $\mathcal{D}$, $\mathcal{Z}$, and $\mathcal{T}$ to denote the corpus with all the document data, the collections of topic distributions of all the documents, and the collections of word distributions of all the topics, respectively.

\paragraph{Notations of Architectures and Learning} \label{sec-learning} With these notations, the task for a topic model is to learn the latent variables of $\mathcal{T}$ and $\mathcal{Z}$ from the observed data $\mathcal{D}$.
More formally, a topic model learns a projection parameterised by $\theta$ from a document's data to its topic distribution: $\vec{z} = \theta(\vec{b})$
and a set of global variables for the word distributions of the topics: $\mathcal{T}$.
In order to learn these parameters, one generates or reconstructs a document's BoW data from its topic distribution, which is modelled by another projection parameterised by $\phi$: $\vec{\tilde{b}} = \phi(\vec{z}, \mathcal{T})$. 
Note that the majority of topic models belong to the category of probabilistic generative models, where $\vec{z}$ and $\vec{b}$ are latent and observed random variables assumed to be generated from certain distributions respectively. The projection from the latent variables to the observed ones is named the generative process, which we further denote as: $\vec{\tilde{b}} \sim p^b_{ \phi}(\vec{z}, \mathcal{T})$ where $\vec{z}$ is sampled from the prior distribution $\vec{z}\sim p^z$. While the inverse projection is named the inference process, denoted as $\vec{z} \sim q^z_{\theta}(\vec{b})$, where $q^z$ is the posterior distribution of $\vec{z}$.
For NTMs, these probabilities are typically parameterised by deep neural networks.

\subsection{Evaluation}
\label{sec-eva}
It is still challenging to comprehensively evaluate and compare the performance of topic models including NTMs. Based on the nature and applications of topic models, the commonly-used metrics are as follows.

\paragraph{Predictive accuracy}
It has been common to measure the log-likelihood of a model on held-out test documents, i.e., the predictive accuracy. 
A more popular metric based on log-likelihood is perplexity, which captures how surprised a model is of new (test) data and is inversely proportional to average log-likelihood per word.
Although log-likelihood or perplexity gives a straight numerical comparison between models, there remain issues: \textbf{1)} As topic models are not for predicting unseen data but learning interpretable topics and representations of seen data, predictive accuracy does not reflect the main use of topic models. 
\textbf{2)} Predictive accuracy does not capture topic quality. Predictive accuracy and human judgement on topic quality are often not correlated~\cite{chang2009reading}, and even sometimes slightly anti-correlated.
\textbf{3)} The estimation of the predictive probability is usually intractable for Bayesian models and different papers may apply different sampling or approximation techniques.
For NTMs, the computation of log-likelihood is even more inconsistent, making it harder to compare the results across different papers. 

\paragraph{Topic Coherence}
Experiments  show  topic coherence (TC) computed with the coherence between a topic's most representative words (e.g, top 10 words) is inline with human evaluation of topic interpretability~\cite{lau2014machine}.
Various formulations have been proposed to compute TC,  we refer readers to~\cite{roder2015exploring} for more details. Most formulations require to compute the general coherence between two words, which are estimated based on word co-occurrence counts in a reference corpus.
Regarding TC: 
\textbf{1)} The ranking of TC scores may vary under different formulations. Therefore, it is encouraged to report TC scores of different formulations or report the average score.
\textbf{2)} The choice of the reference corpus can also affect the TC scores, due to the change of lexical usage, i.e, the shift of word distribution. For example, computing TC for a machine learning paper collection with a tweet dataset as reference may generate inaccurate results. Popular choices of the reference corpus are the target corpus itself or an external corpus such as a large dump of Wikipedia.
\textbf{3)} To exclude less interpretable ``background'' topics, one can select the topics (e.g., top 50\%) with the highest TC and report the average score over those selected topics~\cite{zhao2018dirichlet} or to vary the proportion of the selected topics (e.g, from 10\% to 100\%) and plot TC score at each proportion~\cite{zhao2020neural}.

\paragraph{Topic Diversity}
Topic diversity (TD), as its name implies, measures how diverse the discovered topics are. 
It is preferable that the topics discovered by a model describe different semantic topical meanings. Specifically, \cite{dieng2020topic} defines topic diversity to be the percentage of unique words in the top 25 words.

\paragraph{Downstream Application Performance}
The topic distribution $\vec{z}$ of a document learned by a topic model can be used as the semantic representation of the document, which can be used in document classification, clustering, retrieval, visualisation, and elsewhere.
For document classification, one can train a classification model with the topic distributions learned by a topic model as features and report the classification performance to compare different topic models.
Document clustering can be conducted by two strategies: \textbf{1)} Similar to classification, one can
perform a clustering model (e.g. K-means with different numbers of clusters) on the topic distributions, such as in~\cite{zhao2020neural}; \textbf{2)} Alternatively, topics can be viewed as ``soft'' clusters of documents. Thus, one can use the most significant topic of a document (i.e., the topic with the largest weight in the topic distribution) as the cluster assignment, such as in~\cite{nguyen2015improving}.
For document retrieval, we can use the distance of the topic distributions of two documents as their semantic distance and report retrieval accuracy as a metric of topic modelling~\cite{Larochelle_Lauly_2012}.
For qualitative analysis, a popular choice is to use visualisation techniques (e.g., t-SNE) on $\vec{z}$.

\begin{figure*}[t]
\centering
\begin{tikzpicture}[scale=0.45, every node/.style={transform shape},grow'=down,level distance=1.5in,sibling distance=.05in]
\tikzset{edge from parent/.style= 
            {thick, draw, edge from parent fork down},
         every tree node/.style=
            {minimum width=0.001in,text width=3.6in,align=center}}
\Large            
\Tree
 [
.{\textbf{\LARGE{Frameworks of NTMs}}}
[.\node[draw,rectangle,text width=1in]{Other Frameworks (Section~\ref{sec-other})\\\cite{cao2015novel,Chen_Zaki_2017,peng2018neural,Gui_Leng_Pergola_Zhou_Xu_He_2019,Nan_Ding_Nallapati_Xiang_2019,zhao2020neural}}; ]
[.\node[draw,rectangle,text width=1in]{GNN-NTMs (Section~\ref{sec-gnn})\\\cite{zhu2018graphbtm,Yang_Wu_Gu_Wang_Cao_Jin_Guo_2020,Zhou_Hu_Wang_2020}}; ]
[.\node[draw,rectangle,text width=1in]{GAN-NTMs (Section~\ref{sec-gan})\\\cite{Wang_Zhou_He_2019,Wang_Hu_Zhou_He_Xiong_Ye_Xu_2020,Hu_Wang_Zhou_Xiong_2020}}; ]
[.\node[draw,rectangle,text width=1in]{DocNADE-NTMs (Section~\ref{sec-docnade-ntm})\\\cite{Larochelle_Lauly_2012,gupta2019document,gupta2020neural,gupta2018texttovec}};]
[.\node[draw,rectangle,text width=1in]{VAE-NTMs (Section~\ref{sec-vae})};
[.\node[draw,rectangle,text width=1in]{NTMs with Pre-trained Language Models (Section~\ref{sec-vae-pre})\\\cite{Bianchi_Terragni_Hovy_2020,Thompson_Mimno_2020,chaudhary2020topicbert,Hoyle_Goel_Resnik_2020}}; ]
[.\node[draw,rectangle,text width=1in]{Sequential NTMs (Section~\ref{sec-vae-seq})\\\cite{Nallapati_Melnyk_Kumar_Zhou_2017,Zaheer_Ahmed_Smola,Dieng_Wang_Gao_Paisley_2017,Panwar_Shailabh_Aggarwal_Krishnamurthy_2020,Rezaee_Ferraro_2020}}; ]
[.\node[draw,rectangle,text width=1in]{NTMs for Short Text (Section~\ref{sec-vae-short})\\\cite{Zeng_Li_Song_Gao_Lyu_King_2018,zhu2018graphbtm,Lin_Jiang_Rao_2020,Feng_Zhang_Ding_Rao_Xie_2020,Wu_Li_Zhu_Miao_2020,He_Zhang_Jin_Wang_Dang_Li_2018}}; ]
[.\node[draw,rectangle,text width=1in]{NTMs with Meta-data (Section~\ref{sec-vae-meta})\\\cite{Card_Tan_Smith_2018,Korshunova_Xiong_Fedoryszak_Theis_2020,Wang_Yang,bai2018neural}}; ]
[.\node[draw,rectangle,text width=1in]{Correlated and Structured Topics (Section~\ref{sec-vae-cor})\\\cite{Liu_Huang_Gao_Zhang_Wei_2019,isonuma2020tree,zhang2018whai,esmaeili2019structured}}; ]
[.\node[draw,rectangle,text width=1in]{Variants of Word Distributions (Section~\ref{sec-vae-vd})\\\cite{Jung_Choi_2017,dieng2020topic,Ding_Nallapati_Xiang_2018}}; ]
[.\node[draw,rectangle,text width=1in]{Variants of BoW Data Distributions (Section~\ref{sec-vae-vd})\\\cite{Zhao_Rai_Du_Buntine_Phung_Zhou_2020}}; ]
[.\node[draw,rectangle,text width=1in]{Nonparametric priors for Topics (Section~\ref{sec-vae-vd})\\\cite{nalisnick2016stick,ning2020nonparametric,Miao_Grefenstette_Blunsom_2018,Wu_Rao_Zhang_Xie_Li_Wang_Chen_2020}}; ]
[.\node[draw,rectangle,text width=1.25in]{Variants of Priors for Topics (Section~\ref{sec-vae-vd})\\ \cite{Srivastava_Sutton_2017,zhang2018whai,joo2020dirichlet,Burkhardt_Kramer,Tian_Mao_Zhang_2020,Miao_Grefenstette_Blunsom_2018,Silveira_Carvalho_Cristo_Moens_2018,Lin_Hu_Guo_2019}}; ]]
]
]]   
\end{tikzpicture}
\caption{A taxonomy of the papers regarding neural topic models}
\label{fig-taxonomy}
\end{figure*}
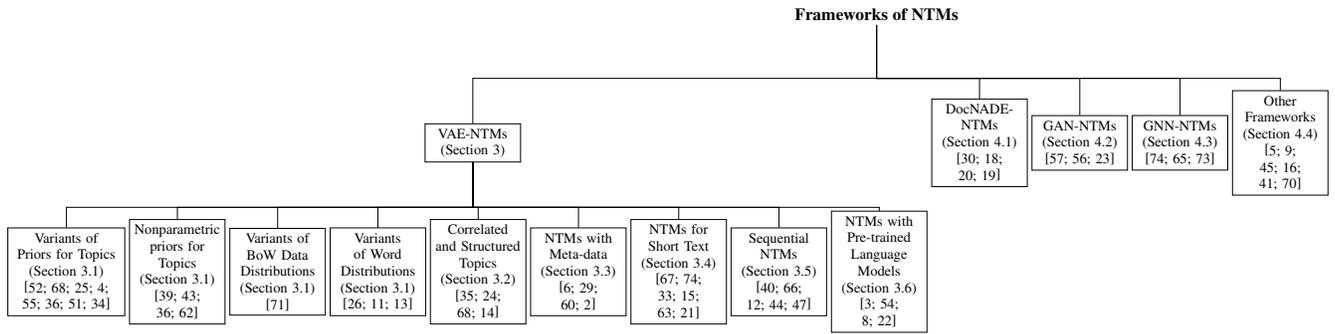

\section{Neural Topic Models with Amortised Variational Inference}
\label{sec-vae}
The recent success of deep generative models such as variational autoencoders (VAEs) and amortised variational inference (AVI)~\cite{kingma2013auto,rezende2014stochastic} has shed light on extending the generative process and amortising the inference process of BPTMs, which is the most popular framework for NTMs. We name this series of models VAE-NTMs.
The basic framework of a VAE follows the description in Section~\ref{sec-learning}, where $\vec{b}$ and $\vec{z}$ are the observed and latent variables respectively and the generative and inference processes are modelled by the DNN-based decoder and encoder respectively.
Following~\cite{kingma2013auto,rezende2014stochastic}, one can learn a VAE model by 
maximising the Evidence Lower BOund (ELBO) of the marginal likelihood of the BoW data $\vec{b}$ in terms of $\theta$, $\phi$, and $\mathcal{T}$:
$\exptt{\vec{z} \sim q^z}{\log \pprob{\vec{b} | \vec{z}}} - \kl{q^z}{p^z},$
where the RHS term is the Kullback-Leiber (KL) divergence.
To compute/estimate gradients, tricks like reparameterisations are usually used to back-propagate gradients through the expectation in the LHS term and approximations are applied when  the analytical form of the KL divergence is unavailable. 

To adapt the VAE framework for topic modelling, there are two key questions to be answered:
\textbf{1)} Different from other applications, the input data of topic modelling has its unique properties, i.e., $\vec{b}$ is a high-dimensional, sparse, count-valued vector and $\vec{s}$ is a variable-length sequential data. How to deal with such data is the first question for designing a VAE topic model. 
\textbf{2)} Interpretability of topics is extremely important in topic modelling. When it comes to a VAE model, how to explicitly or implicitly incorporate the word distributions of topics (i.e., $\mathcal{T}$) to interpret the latent representations or each dimension remains another question.
\cite{Miao_Yu_Blunsom_2016} proposes the first answers to the above questions, where the decoder is developed by specifying the data distribution $p^b$ as: $p^b \eqdef \text{Multi}\left(\softmax{\matr{T}^{T}\vec{z}+ \vec{c}}\right)$.
Here $\vec{z} \in \mathbb{R}^K$ models the topic distribution of a document, $\matr{T} \in \mathbb{R}^{K \times V}$ models the words distributions of the topics, and $\vec{c} \in \mathbb{R}^V$ is the bias. That is to say, $\phi \eqdef \{\vec{c}\}$\footnote{With a slight abuse of notation, we use $\theta$ and $\phi$ to denote the projections or the parameters of the projections.} and $\mathcal{T} \eqdef \{\matr{T}\}$.
For the encoder which takes $\vec{b}$ as input and outputs (the samples of) $\vec{z}$, the paper follows the original VAE:  $p^z \eqdef \mathcal{N}(\vec{0}, \text{diag}_K(\vec{1}))$; $q^z \eqdef \mathcal{N}(\vec{\mu}, \text{diag}_K{(\vec{\sigma}^2)})$, where $\vec{\pi} = \theta_0(\vec{b})$, $\vec{\mu} = \theta_1(\vec{\pi})$, and $\log\vec{\sigma} = \theta_2(\vec{\pi})$. Here, $\theta \eqdef \{\theta_0,\theta_1, \theta_2\}$, all of which are multi-layer perceptrons (MLPs).
To better address the above questions, various configurations of the prior distribution $p^z$, data distribution $p^b$, posterior distribution $q^z$, as well as different architectures of the decoder $\phi$, encoder $\theta$, word distributions of the topics $\mathcal{T}$, have been proposed for VAE-NTMs. Figure~\ref{fig-taxonomy} shows the taxonomy of VAE-NTMs.

\subsection{Variants of Distributions}
\label{sec-vae-vd}
Given the knowledge and experience of BPTMs, $\vec{z}$'s prior plays an important role in the quality of topics and document representations in topic models.
Thus, various constructions of the prior distributions and their corresponding posterior distributions have been proposed for VAE-NTMs, aiming to be better alternatives to the normal distributions used in the original models.

\paragraph{Variants of Prior Distributions for $\vec{z}$.} 
Note that the application of Dirichlet is one of the key successes of LDA for encouraging topic smoothness and sparsity. For VAE-NTMs, one can apply: $p^z \eqdef \text{Dir}(\alpha_0)$ and $q^z \eqdef \text{Dir}(\theta(\vec{b}))$.
However, it is difficult to develop an effective reparameterisation function (RF) for Dirichlet, making it hard to compute the gradient of the expectation in ELBO.
Therefore, various approximations have been proposed.
For example, \cite{Srivastava_Sutton_2017} uses the Laplace approximation, where Dirichlet samples are approximated by these sampled from a logistic normal distribution, whose mean and co-variance are specifically configured.
Recall that the Dirichlet distribution can be simulated by normalising gamma variables, which still do not have non-central differentiable RF but are easier to approximate. Several works have been proposed in this line, such as using the Weibull distribution as the approximation of gamma in~\cite{zhang2018whai}, approximating the cumulative distribution function of gamma with an auxiliary uniform variable in~\cite{joo2020dirichlet}, and leveraging the proposal function of a rejection sampler of the gamma distribution as the RF in~\cite{Burkhardt_Kramer}.
Recently, \cite{Tian_Mao_Zhang_2020} proposes to tackle this challenge by using the so-called rounded RF, which approximates Dirichlet samples by those drawn from the rounded posterior distribution.
In addition to the above methods that are specific to topic modelling or VAEs, other general approaches for distributions without RF can also be used in VAE-NTMs, such as those in~\cite{ruiz2016generalized,naesseth2017reparameterization}. Other than Dirichlet, \cite{Miao_Grefenstette_Blunsom_2018} introduces a Gaussian softmax (GSM) function in the encoder: $q^z \eqdef \softmax{\mathcal{N}(\vec{\mu}, \text{diag}_K{(\vec{\sigma}^2)})}$ and \cite{Silveira_Carvalho_Cristo_Moens_2018} proposes to use a logistic-normal mixture distribution for the prior of $\vec{z}$.
To further enhance the sparsity in $\vec{z}$, \cite{Lin_Hu_Guo_2019} introduces to use the sparsemax function to replace the softmax in GSM.

\paragraph{Nonparametric Prior for $\vec{z}$.} Bayesian Nonparametrics such as the Dirichlet processes and Indian Buffet Processes have been successfully applied in Bayesian topic modelling, enabling to automatically infer the number of topics (i.e., $K$).
As a flexible construction of Dirichlet processes, the stick-breaking process (SBP) is able to generate probability vectors with infinite dimensions, which has been used to the prior of $\vec{z}$ in VAE-NTMs.
Given $\vec{z} \sim \text{SBP}(\alpha_0)$, we have $z_1 = v_1$ and $z_k = v_k \prod_{j < k}(1 - v_j)$ for $k > 1$, where $v_k \sim \text{Beta}(1, \alpha_0)$.
This procedure can be viewed as iteratively breaking a length-one stick into multiple ones and the $k^\text{th}$ iteration breaks the stick at the point of $v_k$. Although not for NTMs, \cite{nalisnick2016stick} proposes to use SBP to generate $\vec{z}$ for VAEs, where VI is done by various approximations to the beta distribution of $v_k$ with truncation.
\cite{ning2020nonparametric} adapts this SBP construction for VAE-NTMs and also proposes to impose an SBP on the corpus level, which serves as the prior for the document-level SBP, forming into a hierarchical model.
In~\cite{Miao_Grefenstette_Blunsom_2018}, the break points $v_k$ are generated from a posterior modelled by a recurrent neural network (RNN) with normal noises as input, making the model able to automatically infer $K$ in a truncation-free manner.
Recently, \cite{Wu_Rao_Zhang_Xie_Li_Wang_Chen_2020} uses the truncated (gamma) negative binomial process to generate discrete vectors for $\vec{z}$ (i.e. each entry of $\vec{z}$ is equivalently generated by an independent Poisson distribution), which gives the model certain ability to be nonparametric.

\paragraph{Variants of Data Distribution $p^b$.}
In addition to imposing different distributions on $\vec{z}$, \cite{Zhao_Rai_Du_Buntine_Phung_Zhou_2020} proposes to replace the multinomial data distribution used in other NTMs with the negative-binomial distribution to capture overdispersion: $\vec{b} \sim \text{NB}(\phi_0(\vec{z}), \phi_1(\vec{z}))$, where two separate decoders $\phi_0$ and $\phi_1$ are proposed to generate the two parameters of the negative-binomial distribution from $\vec{z}$.

\paragraph{Variants of Word Distributions $\mathcal{T}$.}
Conventionally, the collection of the word distributions of the topics $\mathcal{T}$ is a $K \times V$ matrix, i.e., $\matr{T} \in \mathbb{R}^{K \times V}$ with $KV$ free parameters to learn.
In NTMs, it has been popular to factorise the matrix into a product of topic and word embeddings, meaning that the relevance between a topic and a word is captured by their distance in the embedding space.
This construction has been studied in details in~\cite{Jung_Choi_2017,dieng2020topic,Ding_Nallapati_Xiang_2018}.

\subsection{Correlated and Structured Topics}
\label{sec-vae-cor}
Topics discovered by conventional topic models like LDA are usually independent.
An important research direction is to explicitly capture topic correlations (e.g. pairwise relations between topics) or structures (e.g. tree structures of topics), which has been studied in NTMs as well.
Following the framework of VAE with Householder flow, which enables to draw $\vec{z}$ from the normal posterior with a non-diagonal covariance matrix, \cite{Liu_Huang_Gao_Zhang_Wei_2019} develops a more efficient centralised transformation flow for NTMs, which is able to discover pairwise topic correlations by the covariance matrix.
In terms of tree-structured topics, \cite{isonuma2020tree} introduces to generate a series of topics from the root to the leaf of a topic tree with a doubly-recurrent neural network~\cite{alvarez2016tree}.
When applied in topic modelling, the gamma belief network (GBN) can be viewed as a Bayesian model that also discovers three-structured topics, whose inference is done by Gibbs sampling.
\cite{zhang2018whai} introduces the NTM counterpart of GBN, which leverages AVI as the inference process and significantly improves the test time of GBN.
\cite{esmaeili2019structured} proposes an structured VAE-NTM that discovers topics with respect to different aspects, specialising in modelling user reviews.

\subsection{NTMs with Meta-data}
\label{sec-vae-meta}
Conventionally, topic models learn from documents in an unsupervised way.
However, documents are usually associated with rich sets of meta-data on both document and word levels, such as document labels, authorships, and pre-trained word embeddings, which can be used to improve topic quality or document representation quality for supervised tasks (e.g., accuracy of predicting document meta-data).
\cite{Card_Tan_Smith_2018} proposes a VAE-NTM that is able to incorporate various kinds of meta-data, where the BoW data $\vec{b}$ of a document and its labels (e.g., sentiment) are generated with a joint process conditioned on the document's covariates (e.g., publication year) in the decoder and the encoder generates $\vec{z}$ by conditioning on all types of data of the document: BoW, covariates, and labels.
Instead of specifying the generative model as a directed network as in most of topic models, \cite{Korshunova_Xiong_Fedoryszak_Theis_2020} introduces the logistic LDA model whose generative process can be viewed as an undirected graph. In addition to the BoW data, a document's label is also an observed variable in the graph. 
Following a few assumptions of factorisation in the generative process, the paper manually specifies the complete conditional distributions in the graph with the interactions between the latent variables captured by neural networks. The inference is done by the mean-field VI and $\vec{z}$ in the model is further trained to be more discriminative for the classification of labels.
Given a set of documents with labels, \cite{Wang_Yang} uses a VAE-NTM to model a document's BoW data and an RNN classifier to predict a document's label based on its sequential data in a joint training process.
The paper combines the two models by introducing an attention mechanism in the RNN which takes documents' topics into account.
\cite{bai2018neural} proposes to incorporate relational graphs (e.g. citation graph) of documents into NTMs, where the topic distributions of two document are fed into a network with MLPs to predict whether they should be connected.

\subsection{NTMs for Short Texts}
\label{sec-vae-short}
Texts generated on the internet (e.g., tweets, news headlines and product reviews) can be short, meaning that each individual document contains insufficient word co-occurrence information. This results in degraded performance for both BPTMs and NTMs.
To tackle this issue, one can limit a model's capacity and to enhance the contextual information of short texts.
\cite{Zeng_Li_Song_Gao_Lyu_King_2018} proposes a combination of an NTM and a memory network for short text classification in a similar spirit to~\cite{Wang_Yang}.
The main difference is the memory network instead of RNN is responsible for classification, which is informed by the topic distributions learned by the NTM.
To enhance the contextual information of short documents, \cite{zhu2018graphbtm} proposes an NTM whose encoder is a graph neural network (GNN) taking the biterms graph of the words in sampled documents as inputs and outputting the topic distribution for the whole corpus. The model also learns a decoder that reconstructs the input biterms graph.
Despite the novel idea, the model might not be able to generate the topic distribution of an individual document.
To limit a short document to focus on several salient topics, \cite{Lin_Jiang_Rao_2020} introduces to use the Archimedean copulas to regularise the discreteness of topic distributions for short texts.
\cite{Feng_Zhang_Ding_Rao_Xie_2020} proposes an NTM with reinforced content by limiting the number of the active topics for each short document and informing the word distributions of the topics by using pretrained word embeddings.
\cite{Wu_Li_Zhu_Miao_2020} introduces an NTM with vector quantisation over $\vec{z}$, i.e., a document's topic distribution can only be one vector in the learned dictionary in the vector quantisation process.
In addition to maximising the likelihood of the input documents, the paper introduces to minimise the likelihood of the negatively-sampled ``fake documents''.
Although not directly addressing the short text problem for topic modelling, \cite{He_Zhang_Jin_Wang_Dang_Li_2018} introduces NTMs for modelling microblog conversations, by leveraging their unique meta data and structures.

\subsection{Sequential NTMs}
\label{sec-vae-seq}
The flexibility of VAE-NTMs enables to leverage various neural network architectures for the encoder and decoder.
With the help of sequential networks like RNNs, unlike other NTMs working with BoW data (i.e., $\vec{b}$), sequential NTMs (SNTMs) usually take sequences of words of documents (i.e., $\vec{s}$) and are able to capture the orders of words, sentences, and topics.
\cite{Nallapati_Melnyk_Kumar_Zhou_2017} proposes an SNTM working with $\vec{s}$, which samples a topic for each sentence of an input document according to $\vec{z}$ and then generates the word sequence of the sentence with an RNN conditioned on the sentence's topic. Note that $\vec{z}$ is attached to a document and shared across all its sentences.
In~\cite{Zaheer_Ahmed_Smola}, given $\vec{s}$, a word's topic is conditioned on its previous word's and this order dependency is captured by a long short-term memory (LSTM) model.
At the similar period of time, \cite{Dieng_Wang_Gao_Paisley_2017} independently proposes an SNTM whose generative process is similar to~\cite{Zaheer_Ahmed_Smola}, with an additional variable modelling stop words and several variants in the inference process.
Recently, \cite{Panwar_Shailabh_Aggarwal_Krishnamurthy_2020} proposes to use an LSTM with attentions as the encoder taking $\vec{s}$ as input, where the attention incorporates topical information with a context vector that is constructed by topic embeddings and document embeddings.
\cite{Rezaee_Ferraro_2020} introduces an SNTM that is related to \cite{Dieng_Wang_Gao_Paisley_2017}, where instead of marginalising out the discrete topic assignments, the paper proposes to generate them from an RNN model. This helps to avoid using reparameterisation tricks in the variational inference.

\subsection{NTMs with Pre-trained Language Models}
\label{sec-vae-pre}
Recently, pre-trained transformer-based language models such as BERT are becoming ubiquitous in NLP.
Pre-trained on large corpora, such models usually have a fine-grained ability to capture aspects of linguistic context, which can be partially represented by contextual word embeddings.
These contextual word embeddings can provide richer context information than BoW or sequential data, which has been recently used to assist the training of topic models.
Instead of using the BoW or sequential data of a document as the input of the encoder, \cite{Bianchi_Terragni_Hovy_2020} proposes to use the document embedding vector generated by Sentence-BERT~\cite{reimers2019sentence} and to keep the remaining part of an NTM the same as~\cite{Srivastava_Sutton_2017}.
\cite{Thompson_Mimno_2020} shows that the clusters obtained by performing clustering algorithms (e.g., Kmeans) on the contextual word embeddings generated by various pre-trained models can be interpreted as topics, similar to those discovered by LDA.
Having similar ideas with~\cite{Zeng_Li_Song_Gao_Lyu_King_2018,Wang_Yang}, \cite{chaudhary2020topicbert} proposes to combine an NTM with a fine-tuned BERT model by concatenating the topic distribution and the learned BERT embedding of a document as the features for document classification.
\cite{Hoyle_Goel_Resnik_2020} proposes an NTM learned by distilling knowledge from a pre-trained BERT model. Specifically, given a document, the BERT model generates the predicted probability for each word then the paper introduces to average those probabilities to generate a pseudo BoW vector for the document. An NTM following~\cite{Card_Tan_Smith_2018} is used to reconstruct both the actual and pseudo BoW data.

\section{NTMs based on Other Frameworks}
\label{sec-other-framework}
Besides VAE-NTMs, there are other frameworks for NTMs that also draw research attention.
\subsection{NTMs based on Autoregressive Models}
\label{sec-docnade-ntm}
VAE-NTMs gained popularity after VAEs were invented. Before that, NTMs based on the autoregressive framework had been studied.
Specifically, \cite{Larochelle_Lauly_2012} proposes an autoregressive NTM, named DocNADE, similar to the spirit of RNNs, where the predictive probability of a word in a document is conditioned on its hidden state, which is further conditioned on the previous words. A hidden unit can be interpreted as a topic and a document's hidden states capture its topic distribution. The learning is done by maximising the likelihood of the input documents.
Recently, \cite{gupta2019document} extends DocNADE by introducing a structure similar to the bi-directional RNN, which allows to model bi-directional dependencies between words.
\cite{gupta2018texttovec} combines DocNADE with an LSTM for incorporating external knowledge.
\cite{gupta2020neural} extends DocNADE into the life long learning settings.

\subsection{NTMs based on Generative Adversarial Nets}
\label{sec-gan}
Besides VAEs, generative adversarial networks (GANs) are another popular series of deep generative models.
Recently, there are a few attempts on adapting the GAN framework for topic modelling.
\cite{Wang_Zhou_He_2019} proposes a GAN generator that takes a random sample of the Dirichlet distribution as a topic distribution $\tilde{\vec{z}}$ and generates the word distributions of a ``fake'' document conditioning on $\tilde{\vec{z}}$.
A discriminator is introduced to distinguish between generated word distributions and real word distributions obtained by normalising the TF-IDF vectors of real documents.
Although the proposed model is able to discover interpretable topics, it cannot learn topic distributions for documents.
To address this issue, \cite{Wang_Hu_Zhou_He_Xiong_Ye_Xu_2020} introduces an additional encoder that learns $\vec{z}$ for a given document.
Moreover, $\vec{z}$ is concatenated with the word distribution of a document as a real datum and $\tilde{\vec{z}}$ is concatenated with the generated word distribution as a fake datum.
The discriminator is designed to distinguish between the real and fake ones.
\cite{Hu_Wang_Zhou_Xiong_2020} further extends the above model with a CycleGAN framework.

\subsection{NTMs based on Graph Neural Networks}
\label{sec-gnn}
Instead of viewing a document as a sequence or bag of words, one can consider the graph presentations of a corpus of documents.
This perspective enables leveraging a variety of GNNs to discover latent topics.
As discussed in Section~\ref{sec-vae-short}, \cite{zhu2018graphbtm} views a collection of documents as a biterm word graph.
While~\cite{Yang_Wu_Gu_Wang_Cao_Jin_Guo_2020,Zhou_Hu_Wang_2020} model a corpus by a bipartite graph with documents and words as two separate parties and connected by the occurrences of words in documents.
For the former, it directly uses the word occurrences of documents as the weights of the connections between them and for the latter, it uses TF-IDF values instead.

\subsection{NTMs based on Other Frameworks}
\label{sec-other}
In addition to the above frameworks, other kinds of NTMs have also been developed.
An NTM is developed in \cite{cao2015novel} that takes n-gram embeddings (obtained from word embeddings) and a document index as input and then predicts whether an n-gram is in the document.
\cite{Chen_Zaki_2017} proposes an autoencoder model for NTMs where the neurons in the hidden layer of the autoencoder compete with each other, focusing them to be specialised in recognising specific data patterns.
\cite{peng2018neural} proposes an NTM based on matrix factorisation.
\cite{Gui_Leng_Pergola_Zhou_Xu_He_2019} proposes a reinforcement learning framework for NTMs, where the encoder and decoder of an NTM are kept. In addition, an agent takes actions to select the topical-coherent words from a document and uses the selected words as the input document for the encoder. The reward to the agent is the topic coherence of the reconstructed document from the decoder.
\cite{Nan_Ding_Nallapati_Xiang_2019} adapts the framework of Wasserstein auto-encoders (WAEs), which minimises the Wasserstein distance between reconstructed documents from the decoder and real documents, similarly to VAE-NTMs.
\cite{zhao2020neural} recently introduces a NTM based on optimal transport, which directly minimises the optimal transport distance between the topic distribution learned by an encoder and the word distribution of a document.

\section{Applications of NTMs}
\label{sec-application}
Although just recently developed, NTMs have been actively used in various applications. Compared with conventional topic models, NTMs have the appealing advantages of flexibility: \textbf{1)} NTMs are flexible in representing topic distributions of documents and word distributions of topics with either probability vectors or embeddings, and are more easily incorporated into broader models. \textbf{2)} The inference process of NTMs can usually be formulated as an optimisation process with gradients, which is more conveniently integrated with other DNN models for joint training.

Many DNN models used for language such as RNNs, transformers, and attention might not be able to capture long-range dependency well. On the contrary, working on BoW data, NTMs are good at learning  global semantic representations for long texts, which can serve complementary information to the above models.
This leads to a wide range of applications of NTMs in NLP such as language models~\cite{Lau_Baldwin_Cohn_2017,Wang_Gan_Wang_Shen_Huang_Ping_Satheesh_Carin_2018,Xiao_Zhao_Wang_2018,guo2020recurrent,Kawamae_2019}, text generation~\cite{Tang_Li_Jin_2019,Wang_Gan_Xu_Zhang_Wang_Shen_Chen_Carin_2019}, and summarisation~\cite{Cui_Hu_Liu_2020,Zheng_Zhang_Wang_Fan_2020,wang2020friendly}. Due to the space limit of the references, a detailed list of application papers are omitted.

\section{Discussion}
\label{sec-discuss}
In this paper, we have reviewed neural topic models, the most popular research trend of topic modelling in the deep learning era.
A variety of NTMs based on different frameworks have been developed and due to the appealing flexibility, effectiveness, and efficiency, NTMs show a promising potential in a range of applications.
In addition to providing an overview of existing approaches of NTMs, we would like to discuss the following challenges and opportunities.
\textbf{1) Better evaluation: }
As stated in Section~\ref{sec-eva}, evaluation of topic models is challenging. This is mainly because there has not been a unified system of evaluation metrics, making the comparisons across different NTMs harder due to the variety of frameworks, architectures and datasets. For example, VAE-NTMs calculate perplexity using the ELBO, attached to the models with variational inference, which cannot be compared with models without ELBO. Also for topic coherence and downstream performance, the evaluation processes, metrics, settings usually vary in different papers.
As a topic model should be evaluated with comprehensive metrics, it could be tendentious to only use one kind of metric (e.g., topic coherence), which can reflect just one aspect of a model. Therefore, unified platforms and benchmarks for NTMs are needed.
\textbf{2) Richer architectures and applications: } Compared to BPTMs, NTMs offer better flexibility for representing topic distributions for documents and word distributions for topics. Particularly, projecting documents, topics, and words into a unified embedding space transforms the thinking of the relationships between the three.
Given this flexibility, NTMs are expected to get integrated with the most recent neural architectures and play a unique role in richer applications.
\textbf{3) More external knowledge: } With the development of topic models including NTMs, people have not stopped seeking to leverage external knowledge to help the learning, from document meta-data to pre-trained word embeddings. Recently-proposed pre-trained language models (e.g., BERT) provide more advanced, finer-grained, and higher-level representations of semantic knowledge (e.g., contextual word embeddings over global embeddings), which can be leveraged in NTMs to boost performance. Although the marriage between NTMs and language models is still an emerging area, we expect to see more developments in this important direction.

\bibliographystyle{abbrv}
\bibliography{ntm_survey}
\end{document}